\documentclass[11pt,a4paper]{article}
\usepackage{authblk}
\usepackage[hyperref]{acl2019}
\usepackage{times}
\usepackage{latexsym}
\usepackage{multirow}
\usepackage{url}
\usepackage{amsmath}
\usepackage{booktabs}
\usepackage{graphicx}
\usepackage{bbold}

\aclfinalcopy 

\DeclareMathOperator*{\argmax}{arg\,max}
\DeclareMathOperator{\softmax}{softmax}

\title{Probing Neural Network Comprehension of Natural Language Arguments}

\author{\textbf{Timothy Niven}}
\author{\textbf{Hung-Yu Kao}}
\affil{Intelligent Knowledge Management Lab \\
  Department of Computer Science and Information Engineering \\
  National Cheng Kung University \\
  Tainan, Taiwan \\
  \texttt{tim.niven.public@gmail.com}, \texttt{hykao@mail.ncku.edu.tw}}

\date{}

\begin{document}
\maketitle
\begin{abstract}
  We are surprised to find that BERT's peak performance of 77\% on the Argument Reasoning Comprehension Task reaches just three points below the average untrained human baseline. However, we show that this result is entirely accounted for by exploitation of spurious statistical cues in the dataset. We analyze the nature of these cues and demonstrate that a range of models all exploit them. This analysis informs the construction of an adversarial dataset on which all models achieve random accuracy. Our adversarial dataset provides a more robust assessment of argument comprehension and should be adopted as the standard in future work.
\end{abstract}

\noindent
\textbf{This is an updated version of our ACL paper.}

\section{Introduction}

Argumentation mining is the task of determining argumentative structure in natural language text - e.g., which text segments represent claims, and which comprise reasons that support or attack those claims \cite{MochalesM11, LippiT16}. This is a challenging task for machine learners, as it can be hard even for humans to determine when two text segments stand in argumentative relation, as evidenced by studies on argument annotation \cite{HabernalEG14}.

One approach to this problem is to focus on \textit{warrants} \cite{Toulmin58} - a form of world knowledge that permit inferences. Consider a simple argument: ``(1) It is raining; therefore (2) you should take an umbrella.''\footnote{This example adapted from \citeauthor{BlackH12} \shortcite{BlackH12}} The warrant ``(3) it is bad to get wet'' could license this inference. Knowing (3) facilitates drawing the inferential connection between (1) and (2). However it would be hard to find it stated anywhere since warrants are most often left implicit \cite{Walton05}. Thus, on this approach, machine learners must not only reason with warrants but also discover them.

\begin{figure}[t]
\begin{center}
\small

\begin{tabular}{p{0.08\textwidth}p{0.36\textwidth}}
\textbf{Claim} & Google is not a harmful monopoly \\
\textbf{Reason} & People can choose not to use Google \\
\textbf{Warrant} & Other search engines don't redirect to Google \\
\textbf{Alternative} & All other search engines redirect to Google \\
\end{tabular} \\
\vspace{8pt}
\textbf{Reason} (and since) \textbf{Warrant} $\rightarrow$ \textbf{Claim} \\
\textbf{Reason} (but since) \textbf{Alternative} $\rightarrow \lnot$ \textbf{Claim}

\end{center}
\caption{An example of a data point from the ARCT test set and how it should be read. The inference from $R$ and $A$ to $\lnot C$ is by design.}
\end{figure}

The Argument Reasoning Comprehension Task (ARCT) \cite{HabernalWGS17} defers the problem of discovering warrants and focuses on inference. An argument is provided, comprising a claim $C$ and reason $R$. This task is to pick the correct warrant $W$ over a distractor, called the \textit{alternative warrant} $A$. The alternative is written such that $R \land A \rightarrow \lnot C$. An alternative warrant for our earlier example could be ``(4) it is good to get wet,'' in which case we have (1) $\land$ (4) $\rightarrow$ ``($\lnot$2) you shouldn't take an umbrella.'' An example from the dataset is given in Figure 1.

The ARCT SemEval shared task \cite{HabernalWGS18} verified the challenging nature of this problem. Even supplying warrants, learners still need to rely on further world knowledge. For example, to correctly classify the data point in Figure 1 it is at least required to know how consumer choice and web re-directs relate to the concept of monopoly, and that Google is a search engine. All but one participating system in the shared task could not exceed $60\%$ accuracy (on binary classification).

\begin{table*}[t]
\begin{center}
\small
\begin{tabular}{|l|c|ccc|}
\hline
\multirow{2}{*}{} & \multicolumn{1}{c|}{\textbf{Dev}} & \multicolumn{3}{c|}{\textbf{Test}} \\
& Mean & Mean & Median & Max \\
\hline
Human (trained) & & 0.909 $\pm$ 0.11 &  &  \\
Human (untrained) & & 0.798 $\pm$ 0.16 & &  \\
BERT (Large) & 0.701 $\pm$ 0.05 & 0.671 $\pm$ 0.09 & \textbf{0.712} & \textbf{0.770} \\
GIST \cite{ChoiL18} & \textbf{0.716} $\pm$ 0.01 & \textbf{0.711} $\pm$ 0.01 & &  \\
BERT (Base) & 0.680 $\pm$ 0.02 & 0.623 $\pm$ 0.07 & 0.651 & 0.685 \\
World Knowledge \cite{BotschenSG18} & 0.674 $\pm$ 0.01 & 0.568 $\pm$ 0.03 & & 0.610 \\
BoV & 0.639 $\pm$ 0.02 & 0.564 $\pm$ 0.02 & 0.569 & 0.595 \\
BiLSTM & 0.658 $\pm$ 0.01 & 0.552 $\pm$ 0.02 & 0.552 & 0.592 \\
\hline
\end{tabular}
\end{center}
\caption{Baselines and BERT results. Our results come from 20 different random seeds ($\pm$ gives the standard deviation). The mean for BERT Large is skewed by the $5/20$ random seeds for which it failed to train, a problem noted by \citeauthor{DevlinMKK18} \shortcite{DevlinMKK18}. We therefore consider the median a better measure of BERT's average performance. The mean of the non-degenerate runs for BERT (Large) is $0.716 \pm 0.04$.}
\end{table*}

It is therefore surprising that BERT \cite{DevlinMKK18} achieves $77\%$ test set accuracy with its best run (Table 1), only three points below the average (untrained) human baseline. Without supplying the required world knowledge for this task it does not seem reasonable to expect it to perform so well. This motivates the question: what has BERT learned about argument comprehension?

\begin{figure}[t]
\centering
\includegraphics[width=0.4\textwidth]{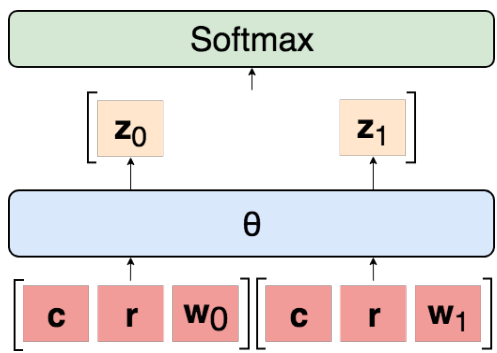}
\caption{General architecture of the models in our experiments. Logits are independently calculated for each argument-warrant pair then concatenated and passed through softmax.}
\end{figure}

To investigate BERT's decision making we looked at data points it finds easy to classify over multiple runs. \citeauthor{HabernalWGS18} \shortcite{HabernalWGS18} performed a similar analysis with the SemEval submissions, and consistent with their results we found that BERT exploits the presence of cue words in the warrant, especially ``not.'' Through probing experiments designed to isolate such effects, we demonstrate in this work that BERT's surprising performance can be entirely accounted for in terms of exploiting spurious statistical cues.

However, we show that the major problem can be eliminated in ARCT. Since $R \land A \rightarrow \lnot C$, we can add a copy of each data point with the claim negated and the  label inverted. This means that the distribution of statistical cues in the warrants will be mirrored over both labels, eliminating the signal. On this adversarial dataset all models perform randomly, with BERT achieving a maximum test set accuracy of $53\%$. The adversarial dataset therefore provides a more robust evaluation of argument comprehension and should be adopted as the standard in future work on this dataset.

\section{Task Description and Baselines}

Let $i = 1,\dots,n$ index each point in the dataset $\mathcal{D}$, where $|\mathcal{D}| = n$. The two candidate warrants in each case are randomly assigned a binary label $j \in \{0, 1\}$, such that each has an equal probability of being correct. The inputs are the representations for the claim $\mathbf{c}^{(i)}$, reason $\mathbf{r}^{(i)}$, warrant zero $\mathbf{w}^{(i)}_0$, and warrant one $\mathbf{w}^{(i)}_1$. The label $y^{(i)}$ is a binary indicator corresponding to the correct warrant.

The general architecture for all models is given in Figure 2. Shared parameters $\boldsymbol{\theta}$ are learned to classify each warrant independently with the argument, yielding the logits:
\begin{equation*}
    z^{(i)}_j = \boldsymbol{\theta} [\boldsymbol{c}^{(i)}; \boldsymbol{r}^{(i)}; \boldsymbol{w}^{(i)}_j]
\end{equation*}
\noindent
These are then concatenated and passed through softmax to determine a probability distribution over the two warrants $\mathbf{p}^{(i)} = \softmax([z^{(i)}_0, z^{(i)}_1])$. The prediction is then $\hat{y}^{(i)} = \argmax_j \mathbf{p}^{(i)}$.

\begin{figure*}[t]
\centering
\includegraphics[width=0.8\textwidth]{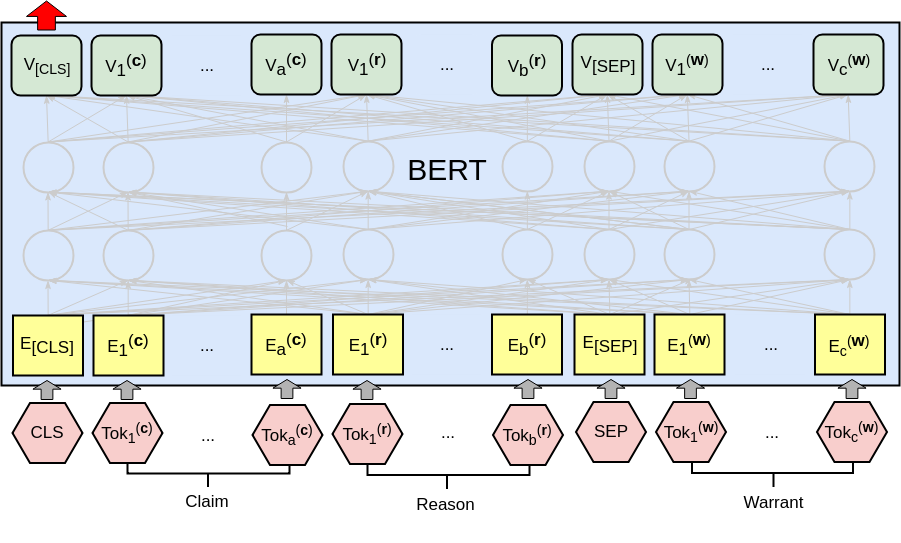}
\caption{Processing an argument-warrant pair with BERT. The reason (with word pieces of length $a$) and claim (length $b$) together form the first utterance, and the warrant (length $c$) is the second. The final CLS vector is then passed to a linear layer to calculate the logit $z^{(i)}_j$.}
\end{figure*}

The baselines are a bag of vectors (BoV),  bidirectional LSTM \cite{HochreiterS97} (BiLSTM), the SemEval winner GIST \cite{ChoiL18}, the best model of \citeauthor{BotschenSG18} \shortcite{BotschenSG18}, and human performance (Table 1). For all of our experiments we use grid search to select hyperparameters, dropout regularization \cite{SrivastavaHKSS14}, and Adam \cite{KingmaB14} for optimization. We anneal the learning rate by $1/10$ when validation accuracy drops. The final parameters come from the epoch with maximum validation accuracy. The BoV and BiLSTM inputs are $300$-dimensional GloVe embeddings trained on $640$B tokens \cite{PenningtonSM14}. Code to reproduce all experiments, and detailing all hyperparameters, is provided on GitHub.\footnote{\hyperlink{https://github.com/IKMLab/arct2.git}{https://github.com/IKMLab/arct2.git}}

\section{BERT}

Our BERT classifier is visualized in Figure 3. The claim and reason are joined to form the first text segment, which is paired with each warrant and independently processed. The final layer CLS vector is passed to a linear layer to obtain the logits $z^{(i)}_j$. The whole architecture is fine-tuned.  The learning rate is $2e^{-5}$ and we allow a maximum of $20$ training epochs, taking the parameters from the epoch with the best validation set accuracy. We use the Hugging Face PyTorch implementation.\footnote{\hyperlink{https://github.com/huggingface/pytorch-pretrained-BERT}{https://github.com/huggingface/pytorch-pretrained-BERT}}

\citeauthor{DevlinMKK18} \shortcite{DevlinMKK18} report that, on small datasets, BERT sometimes fails to train, yielding degenerate results. ARCT is very small with $1,210$ training observations. In $5/20$ runs we encountered this phenomenon, seeing close to random accuracies on validation and test sets. These cases occurred where training accuracy was also not significantly above random ($< 80$\%). Removing the degenerate runs, BERT's mean is $71.6 \pm 0.04$., which would beat the previous state of the art - as would the median of $71.2\%$, which is a better average than the overall mean since it is not skewed by the degenerate cases. However, our main finding is that these results are not meaningful and should be discarded. In the following sections we focus on BERT's peak performance of $77\%$ to make this case.

\section{Statistical Cues}

The major source of spurious statistical cues in ARCT comes from uneven distributions of linguistic artifacts over the warrants, and therefore over the labels. This section aims to demonstrate the presence and nature of these cues. We only consider unigrams and bigrams, although more sophisticated cues may be present. To this end, we aim to calculate how beneficial it is for a model to exploit a cue $k$, and how pervasive it is in the dataset (indicating the strength of the signal).

Formally, let $\mathbb{T}^{(i)}_j$ be the set of tokens in the warrant for data point $i$ with label $j$. We define a cue's \textit{applicability} $\alpha_k$ as the number of data points where it occurs with one label but not the other:
\begin{align*}
    \alpha_k = \sum_{i=1}^n \mathbb{1} \left[ \exists j, k \in \mathbb{T}^{(i)}_j \land k \notin \mathbb{T}^{(i)}_{\lnot j} \right]
\end{align*}

\noindent
The \textit{productivity} $\pi_k$ of a cue is defined as the proportion of applicable data points for which it predicts the correct answer:
\begin{align*}
    \pi_k = \frac{\sum_{i=1}^n \mathbb{1} \left[ \exists j, k \in \mathbb{T}^{(i)}_j \land k \notin \mathbb{T}^{(i)}_{\lnot j} \land y_i = j \right]}{\alpha_k}
\end{align*}

\noindent
Finally, we define the \textit{coverage} $\xi_k$ of a cue as the proportion of applicable cases over the total number of data points: $\xi_k = \alpha_k / n$. In these terms, the productivity of a cue measures the benefit of exploiting it, while coverage measures the strength of the signal it provides. With $m$ labels, if $\pi_k > 1/m$ then the presence of a cue is going to be useful for the task and a machine learner would do well to make use of it. 

\begin{table}[t]
\begin{center}
\small
\begin{tabular}{|l|cc|}
\hline
 & \textbf{Productivity} & \textbf{Coverage} \\
\hline
\textbf{Train} & 0.65 & 0.66 \\
\textbf{Validation} & 0.62 & 0.44 \\
\textbf{Test} & 0.52 & 0.77 \\
\hline
\textbf{All} & \textbf{0.61} & \textbf{0.64} \\
\hline
\end{tabular}
\end{center}
\caption{Productivity and coverage of using the presence of ``not'' in the warrant to predict the label in ARCT. Across the whole dataset, if you pick the warrant with ``not'' you will be right $61\%$ of the time, which covers $64\%$ of all data points.}
\end{table}

The productivity and coverage of the strongest unigram cue we found (``not'') is given in Table 2. It provides a particularly strong training signal. While it is less productive in the test set, it is just one among many such cues. We found a range of other unigrams, albeit with less overall productivity, mostly being high frequency words such as ``is,'' ``do,'' and ``are.'' Bigrams that occurred with not, such as ``will not'' and ``cannot,'' were also found to be highly productive. These statistics indicate the nature of the problem. In the next section we demonstrate that our models are in fact exploiting these cues.

\section{Probing Experiments}

\begin{figure*}[t]
\begin{center}
\begin{tabular}{|l|l|l|}
\hline
& \multicolumn{1}{c|}{\small{\textbf{Original}}} &  \multicolumn{1}{c|}{\small{\textbf{Adversarial}}} \\
\hline
\small{\textbf{Claim}} & \small{Google is not a harmful monopoly} & \small{Google is a harmful monopoly} \\
\small{\textbf{Reason}} & \small{People can choose not to use Google} & \small{People can choose not to use Google} \\
\small{\textbf{Warrant}} & \small{Other search engines do not redirect to Google} & \small{All other search engines redirect to Google} \\
\small{\textbf{Alternative}} & \small{All other search engines redirect to Google} & \small{Other search engines do not redirect to Google} \\
\hline
\end{tabular}
\end{center}
\caption{Original and adversarial data points. The claim is negated and the warrants are swapped. The assignment of labels to $W$ and $A$ are kept the same. By including both, the distribution of linguistic artifacts in the warrants are thereby mirrored around the labels, eliminating the major source of spurious statistical cues in ARCT.}
\end{figure*}

If a model is exploiting distributional cues over the labels, then if trained only on the warrants (W) it should perform relatively well. The same can be said for removing either just the claim, leaving the reason and warrant (R, W), or removing the reason (C, W). The latter setups allow the models to additionally consider cues in the reasons and claims, as well as cues holding over their combinations with the warrants. Each of these setups breaks the  task since we no longer have an argument to match with a warrant.

Experimental results are given in Table 3. On warrants alone (W) BERT achieves a maximum $71\%$ accuracy. That leaves only six percentage points to account for its peak of $77\%$. We find a gain of four percentage points for (R, W) over (W), and a gain of two for (C, W), accounting for the missing six points. Based on this evidence our major finding is that the entirety of BERT's performance can be accounted for in terms of exploiting spurious statistical cues.

\begin{table}[t]
\begin{center}
\small
\begin{tabular}{|l|ccc|}
\hline
\multirow{2}{*}{} & \multicolumn{3}{c|}{\textbf{Test}} \\
& \textbf{Mean} & \textbf{Median} & \textbf{Max} \\
\hline
BERT & \textbf{0.671} $\pm$ 0.09 & \textbf{0.712} & \textbf{0.770} \\
BERT (W) & 0.656 $\pm$ 0.05 & 0.675 & 0.712 \\
BERT (R, W) & 0.600 $\pm$ 0.10 & 0.574 & 0.750 \\
BERT (C, W) & 0.532 $\pm$ 0.09 & 0.503 & 0.732 \\
\hline
BoV & 0.564 $\pm$ 0.02 & 0.569 & 0.595 \\
BoV (W) & 0.567 $\pm$ 0.02 & 0.572 & 0.606 \\
BoV (R, W) & 0.554 $\pm$ 0.02 & 0.557 & 0.579 \\
BoV (C, W) & 0.545 $\pm$ 0.02 & 0.544 & 0.589 \\
\hline
BiLSTM & 0.552 $\pm$ 0.02 & 0.552 & 0.592 \\
BiLSTM (W) & 0.550 $\pm$ 0.02 & 0.547 & 0.577 \\
BiLSTM (R, W) & 0.547 $\pm$ 0.02 & 0.551 & 0.577 \\
BiLSTM (C, W) & 0.552 $\pm$ 0.02 & 0.550 & 0.601 \\
\hline
\end{tabular}
\end{center}
\caption{Results of probing experiments with BERT Large, and the BoV and BiLSTM baselines. These results indicate that BERT's peak $77\%$ performance can be entirely accounted for by exploiting spurious cues. By just considering warrants (W) we can get to $71\%$. Adding cues over reasons (R, W) and claims (C, W) accounts for the remaining six points.}
\end{table}

\section{Adversarial Test Set}

The major problem of statistical cues over warrants in ARCT can be eliminated as a solution, due to the original design of the dataset. Given that $R \land A \rightarrow \lnot C$, we can produce adversarial examples by negating the claim and inverting the label for each data point (Figure 4), which are then combined with the original data. This eliminates the problem by mirroring the distributions of cues around both labels. The negation of most claims in the validation and test sets already exist elsewhere in the dataset. The remaining claims were manually negated by a native English speaker.

We trained on the original data,\footnote{The results reported utilize a training set augmented by adding a copy of each data point with the warrants flipped and the label inverted. Comparable results are achieved with the original data, without this augmentation.} and validated and tested on the adversarial data. The results, given in Table 4, show BERT's peak performance has dropped to $53\%$, with mean and median at $50\%$. The spurious statistics have been eliminated as a solution, as expected.\footnote{Training on the negated data does lead to above random performance, but this is due to exploiting common statistics holding over claims and warrants. Our experimental setup breaks these as solutions by having different distributions over these heuristics in the training and test sets. We report this result in detail in a forthcoming paper.} This result better apts with our intuitions about this task: with little to no understanding about the reality underlying these arguments, good performance shouldn't be feasible.

\section{Related Work}

The most successful previous work on ARCT \cite{ChoiL18, ZhaoLLZY18, NivenK18} involved transfer learning from Natural Language Inference (NLI) datasets \cite{BowmanAPM15, WilliamsNB17}, and utilized effective NLI models such as ESIM \cite{ChenZLWJ16} and InferSent \cite{ConneauKSBB17}. More recently, \citeauthor{BotschenSG18} \shortcite{BotschenSG18} added FrameNet knowledge with modest performance gains. These models should be evaluated on our adversarial dataset. In particular it will be interesting if \citeauthor{BotschenSG18}'s model stands out due to the inclusion of some of the required world knowledge.

There is much recent work focusing on statistical cues in datasets in vision \cite{JoB17} and NLP \cite{SanchezMR18, McCoyPL19, GururanganSLSBS18, GlocknerSG18, PoliakNHRD18, RajpurkarJL18, JiaL17}. Similar to our experiment with warrants, \citeauthor{PoliakNHRD18} \shortcite{PoliakNHRD18} classified NLI data based on the hypothesis only. A similar experiment to our probing task was performed by \citeauthor{NivenK18} \shortcite{NivenK18}, but only with reasons and warrants. They found that independent warrant classification with shared parameters provides some regularization against warrant-label cues. However, this does not solve the problem since the presence of a cue is enough to increase the logits for either warrant.

The original ARCT paper \cite{HabernalWGS17} reported results with a \textit{training} set created in the same way as our adversarial dataset, that also led to random accuracy on the original test set. They suggested it could be that high similarity between the data points made the problem too difficult for the simple models they implemented. Our work indicates the necessity of applying this transformation to the entire dataset in order to obtain a more robust evaluation by eliminating solutions based on spurious statistical cues.

\begin{table}[t]
\begin{center}
\small
\begin{tabular}{|l|ccc|}
\hline
\multirow{2}{*}{} & \multicolumn{3}{c|}{\textbf{Test}} \\
& \textbf{Mean} & \textbf{Median} & \textbf{Max} \\
\hline
BERT & \textbf{0.504} $\pm$ 0.01 & \textbf{0.505} & \textbf{0.533} \\
BERT (W) & 0.501 $\pm$ 0.00 & 0.501 & 0.502 \\
BERT (R, W) & 0.500 $\pm$ 0.00 & 0.500 & 0.502 \\
BERT (C, W) & 0.501 $\pm$ 0.01 & 0.500 & 0.518 \\
\hline
\end{tabular}
\end{center}
\caption{Results for BERT Large on the adversarial test set with adversarial training and validation sets.}
\end{table}

\section{Conclusion}

ARCT provides a fortuitous opportunity to see how stark the problem of exploiting spurious statistics can be. Due to our ability to eliminate the major source of these cues, we were able to show that BERT's maximum performance fell from just three points below the average untrained human baseline to essentially random. To answer our question in the introduction: BERT has learned nothing about argument comprehension. 

However, our investigations confirmed that BERT is indeed a very strong learner. Analysis of easy to classify data points showed reliance on a \textit{lower} proportion of the strongest cue word than the BoV and BiLSTM - i.e. BERT has learned when to ignore the presence of ``not'' and focus on different cues. This indicates an ability to exploit much more subtle joint distributional information. As our learners get stronger, controlling for spurious statistics becomes more important in order to have confidence in their apparent performance. Taken with a growing body of previous work, our results indicate the need for further research into the extent of this problem in NLP more generally.

The adversarial dataset should be adopted as the standard in future work on ARCT. We hope that providing a more robust evaluation will help to spur more productive research on this problem.

\section*{Acknowledgments}

We would like to thank Ivan Habernal, and the reviewers, for their helpful comments.

\bibliography{acl2019}
\bibliographystyle{acl_natbib}

\end{document}